\documentclass[10pt]{article}

\usepackage{fullpage}
\usepackage{setspace}
\usepackage{parskip}
\usepackage{titlesec}
\usepackage[section]{placeins}
\usepackage{xcolor}
\usepackage{breakcites}
\usepackage{lineno}
\usepackage{hyphenat}

\renewenvironment{abstract}
  {{\bfseries\noindent{\abstractname}\par\nobreak}\footnotesize}
  {\bigskip}

\titlespacing{\section}{0pt}{*3}{*1}
\titlespacing{\subsection}{0pt}{*2}{*0.5}
\titlespacing{\subsubsection}{0pt}{*1.5}{0pt}

\usepackage{authblk}

\usepackage{longtable}
\usepackage{pdflscape} 
\usepackage{rotating}
\usepackage{booktabs}  
\usepackage{siunitx}  
\usepackage{graphicx}%
\usepackage{multirow}%
\usepackage{amsmath,amssymb,amsfonts}%
\usepackage{amsthm}%
\usepackage{mathrsfs}%
\usepackage[title]{appendix}%
\usepackage{xcolor}%
\usepackage{textcomp}%
\usepackage{manyfoot}%
\usepackage{booktabs}%
\usepackage{algorithm}%
\usepackage{algorithmicx}%
\usepackage{algpseudocode}%
\usepackage{listings}%
\usepackage{graphicx}
\usepackage{subcaption}
\usepackage{graphicx}
\usepackage{amsmath}
\let\cline\cmidrule
\usepackage{multirow}

\usepackage{adjustbox}
\usepackage{amssymb}
\usepackage{graphicx}
\usepackage[space]{grffile}
\usepackage{latexsym}
\usepackage{textcomp}
\usepackage{longtable}
\usepackage{tabulary}
\usepackage{adjustbox}
\usepackage{amssymb}
\usepackage{pifont}

\usepackage{booktabs,array,multirow}
\usepackage{amsfonts,amsmath,amssymb}
\providecommand\citet{\cite}
\providecommand\citep{\cite}

\newif\iflatexml\latexmlfalse

\usepackage[utf8]{inputenc}
\usepackage[english]{babel}
\usepackage{float}

\begin{document}

\title{\textbf{PDC-ViT : Source Camera Identification using Pixel Difference Convolution and Vision Transformer}}

\author[1]{Omar Elharrouss*}
\author[2]{Younes Akbari}
\author[2]{Noor Almaadeed}
\author[2]{Somaya Al-Maadeed}
\author[3]{Fouad Khelifi}
\author[4]{Ahmed Bouridane}

\affil[1]{Department of Computer Science and Software Engineering, College of Information Technology, United Arab Emirates University.}

\affil[2]{Department of Computer Science and Engineering, Qatar University, Doha, Qatar.}

\affil[3]{Department of Computer and Information Sciences, Northumbria University, Newcastle, UK.}

\affil[4]{Center for Data Analytics and Cybernetics, University of Sharjah, Sharjah, United Arab Emirates.}


\vspace{-1em}

\begingroup
\let\center\flushleft
\let\endcenter\endflushleft
\maketitle
\endgroup

\selectlanguage{english}
\begin{abstract}
Source camera identification has emerged as a vital solution to unlock incidents involving critical cases like terrorism, violence, and other criminal activities. The ability to trace the origin of an image/video can aid law enforcement agencies in gathering evidence and constructing the timeline of events. Moreover, identifying the owner of a certain device narrows down the area of search in a criminal investigation where smartphone devices are involved. This paper proposes a new pixel-based method for source camera identification, integrating Pixel Difference Convolution (PDC) with a Vision Transformer network (ViT), and named PDC-ViT. While the PDC acts as the backbone for feature extraction by exploiting Angular PDC (APDC) and Radial PDC (RPDC). These techniques enhance the capability to capture subtle variations in pixel information, which are crucial for distinguishing between different source cameras. The second part of the methodology focuses on classification, which is based on a Vision Transformer network. Unlike traditional methods that utilize image patches directly for training the classification network, the proposed approach uniquely inputs PDC features into the Vision Transformer network. To demonstrate the effectiveness of the PDC-ViT approach, it has been assessed on five different datasets, which include various image contents and video scenes. The method has also been compared with state-of-the-art source camera identification methods. Experimental results demonstrate the effectiveness and superiority of the proposed system in terms of accuracy and robustness when compared to its competitors. For example, our proposed PDC-ViT has achieved an accuracy of 94.30\%, 84\%, 94.22\% and 92.29\% using the Vision dataset, Daxing dataset, Socrates dataset and QUFVD dataset, respectively. 

\end{abstract}%

\textbf{keywords} Source camera identification, Deep learning method,  Pixel Difference convolution,  Vision Transformers network.
\sloppy

\par\null\selectlanguage{english}


{\label{507127}}

\section{Introduction}
Taking Videos and pictures for different purposes is one of the most repeated habits in our daily lives. From making memories to documenting some events, capturing video clips or taking pictures has become a simple and straightforward action. However, some of these captured sequences can be the witness of many suspicious cases including critical cases such as terrorism, violence, crimes, etc \cite{i1}. The identification of the devices that were used to capture these events can be a very useful tool to unlock and discover the capture events including the owner of these devices thus making the investigation more efficient while minimizing the circle of the suspicious persons. In addition, the shared videos on different social media can also be involved in some investigations, while the devices used can be identified \cite{i2}. This identification can also be part of other tasks like detecting if the video/image is modified/forgotten thus helping with counterfeiting problems \cite{1}.

With the development of computer vision and Deep Learning techniques, the identification of the source camera has become much easier. Many methods have been proposed, and different datasets such as Vision \cite{vision}, Daxing  \cite{daxing}, Socrates \cite{socrates}, QUFVD \cite{QUFVD}, and Video-ACID  \cite{acid}, have been collected to assist in the evaluation of these methods for efficient learning. Source camera identification, which can be represented as a classification problem, has been a hot topic in the last decade. Therefore, exploiting all the proposed networks can simplify the analysis of video/image to identify the source camera. For example, the authors in \cite{2} deployed Inception and Xception networks to design a source camera identification model. Using a simple network of convolutional-pooling layers, the authors in \cite{3} proposed a mobile device identification method based on the noise pattern or camera fingerprint on a dataset of five devices named MICHE-I Dataset. In the same context, but for digital camera identification, the authors in \cite{4} discussed a CNN-based network that exploits kernels of $5\times5$ for convolutional layers and $2\times2$ for pooling layers. Using simple CNN-based models, the proposed paper aims to further improve the performance of the digital camera and phone camera identification. However, identifying the source camera is still challenging because the analysis of these videos/images needs to focus on the representation of pixels (pixel-based treatment) and not on the image's content. In this context, a considerable development has been reached in this stage, but these techniques need to be further improved by considering the new phone brands produced every year as well as the development of camera resolutions for all devices.

In the same context, and in order to analysis of the pixel variations and the regional connections between the pixels (pixel neighboring analysis), we proposed a new source camera identification method based on Pixel Difference Convolution (PDC) and Vision Transformer. Pixel Difference Convolutions are used for feature extraction instead of using popular deep learning networks like ResNet or VGG. Then, the extracted PDC features are used as input for the Vision Transformer network which is used for the classification part. The contributions in this paper are presented as follows:

\begin{itemize}
    \item A deep learning-based system for source camera identification using PDC and a Vision Transformer network.
    
    \item Combined angular and radial pixel difference convolutions for efficient deep feature extraction.
    \item Extensive experiments showing the superiority of the proposed system over state-of-the-art techniques
    
\end{itemize}

The paper is organized as follows. The related works are presented in section 2. Section 3 describes the proposed source camera identification approach. The obtained results and discussion of them are provided in Section 4. While the challenges and future direction are discussed in Section 5. A conclusion is presented in section 6.

\section{Related works}

Nowadays, image and video become the most used techniques for important moments acquisition and memory saving. However, some of these images/videos can be used to capture many other illegal scenarios which makes these techniques a useful alternative for forensic investigations. This not only includes the analysis of the content of the video and images but the identification of the make and model of the device's owner, too. This can be performed by identifying the device used for capturing the videos hence leading to the identification of the device's owner With the diversity of existing devices available in the market, it is very often difficult to identify the device due to the fact that the make (manufacturer) may have several models. For these reasons, the process of identifying a source camera remains a challenging task. Although several methods have been proposed including statistical and machine learning methods, this task can be carried out successfully with creditable performance, especially with the maturity of Deep Machine technology.  This is evidenced by the recent continuous interest and attention to source camera identification from the multimedia forensic community.

To identify the source camera a set of methods and datasets have been proposed and available in the literature using statistical and deep learning-based techniques. These methods can be categorized based on the type of data used in the source camera analyses: Image-based and video-based techniques. While image-based methods work on the analysis of adjacent pixels to identify cameras, video-based approaches are based on the pixel and the temporal relation between the frames of the video. For example, the authors in \cite{5} proposed a source camera identification method using patch-selection and CNN-based methods. From each image, the method selected some small patches for feature extraction using a residual deep-learning model. In the same context, Wang et al. \cite{6} proposed a clustering approach, named EDCO, to classify the source camera. Using a simple CNN-based model, the authors in \cite{i3} replaced the well-known ReLU activation function with a hyperbolic tangent (TanH) function. The same activation function is used in \cite{i4} for implementing the proposed deep-learning mode named ACFM-based. In the same context, and using the similarity between two patches, the authors in \cite{i5} proposed a deep-learning-based model to decide whether the forensic trace of two different patches are the same or not. Using this technique, a source camera identification can be performed based on this similarity computation. Exploiting multi-scale High Pass Filter (HPF), the authors in \cite{i6} proposed a CNN model based on ResNet blocks to identify the camera model. The use of HPFs allows the model to learn from different features of high pass features. The authors in \cite{r1} proposed a frame-based deep learning method for source camera identification named MISLNet, the same architecture, is used in \cite{r2}, which is an extended version of a constrained convolutional layer. The constrained convolutional layer exploited kernels of $5\times5$ instead of $3\times3$ in the normal convolutional layers which make the learning from adjacent pixels without caring out the region content of an image. The first use of this layer in \cite{vision} shows an improvement in recognition accuracy compared with deep learning architectures without this layer.

Pixel-based processing is a common approach to differentiate between two images captured by two different sensors. Considering this principle, some existing methods are focusing on exploiting the pixels as well as the noise in the image to identify the source camera. From these techniques, photo-response non-uniformity (PRNU), which is a widely used approach as a fingerprint for identifying individual imaging devices, has been widely used as feature extraction to implement source camera identification methods. For example, Xiao et al. \cite{r4} proposed a deep learning network to identify source cameras from PRNU features which are used here as pre-processing. DHDN or Densely-connected Hierarchical Denoising Network is inspired from U-Net architecture where the model is trained using two datasets including the Dresden camera dataset and the Daxing smartphone dataset. By analyzing the noise artifacts left on the images as a pre-processing step, the camera model can be identified like in \cite{r5} or to analyze adversarial attacks like in \cite{r6}. Siamese Network architecture is another method for source camera identification proposed by Sameer et al. \cite{r7}.  The proposed method used Multi-Layer Perceptron (MLP) for sharing weights between the two branches of the network. In \cite{r8}, the authors used pre-trained backbones for implementing the source camera identification process using MobileNet and ResNet50 as the backbones of the method. The same networks are exploited for an extended version using Constrained layers using kernels of 5×5 instead of 3×3 in the original ones. This technique was also in various works in \cite{4,r1,r2}. In another method \cite{r9}, the authors move from identification to verification of the source camera which is an important task for unknown models or the models that are not used in the training task. In \cite{r10}, the authors proposed a sequential-based method for source camera identification using PRNU features. In identifying the source camera from a prepared database using the proposed method, a similarity method is used to verify if a current image is similar to one of the existing models in the database. 

\begin{table*}[!t]
\footnotesize
\caption{Summary of related source camera identification methods}
\label{table_summ}
\centering
\begin{tabular}{l|p{6cm}|p{3cm}|p{2cm}}
\hline
\textbf{Method} &\textbf{Technique} & \textbf{Dataset} & \textbf{Type of data} \\

\hline
Ferreira et al. (2018) \cite{2} & Inception and Xception&SPCup& Image \\
\hline
Freire-Obregon et al. (2019) \cite{3}& Camera fingerprint & MICHE-I & Image\\
\hline
Bernacki et al. (2021) \cite{4} & CNN with $5\times5$ kernels &Dresden &Image \\

\hline
Liu et al. (2021) \cite{5}  & Patch selection, CNN with Deep Residual network & Dresden & Image\\
\hline
Wang et al. (2022) \cite{6}  & Data clustering & Dresden & Image\\
\hline
Wang et al. (2018) \cite{i3}&CNN with hyperbolic tangent (TanH)&Socrates, Dresden & Image  \\
\hline
ACFM (2017) \cite{i4}&CNN with hyperbolic tangent (TanH)&Socrates, Dresden & Image \\
\hline
Mayer el al. (2020) \cite{i5}&CNN patch-based similarity &Socrates, Dresden& Image  \\ \hline
Ding et al. (2019) \cite{i6}&CNN, ResNet block, Multi-scale High Pass Filter (HPF) & Socrates, Dresden& Image\\
\hline
Hosler et al. (2019) \cite{r1}& MISLNet with $5\times5$ & Self-collected & Image\\
\hline
Timmerman et al. (2020) \cite{r2} & MISLNet with $5\times5$& \cite{r1} dataset & Image\\\hline
DHDN (2022) \cite{r4} & PRNU features, CNN&Vision, Daxing & Image \\ \hline
Marra et al. (2018) \cite{r6} & DenseNet, XceptionNet & Vision & Image\\\hline
Sameer et al. (2020) \cite{r7}&Siamese Network architecture& Vision, Dresden &Image\\ \hline
Bennabhaktula et al. (2022) \cite{r8}&MobileNe, ResNet50& VISION,  QUFVD&Image\\ \hline
Lawgaly et al. (2022) \cite{r10} & PRNU estimation & Video-ACID &Video\\\hline
Rana et al. (2024) \cite{add1}  & Two-strean of RGB and noise CNN &Dresden, Socrates &  Image    \\\hline
Akbari et al. (2024) \cite{add2}   &  PRNU, CNN     &Daxing, QUFVD     &  Image         \\\hline
Huan et al. (2024) \cite{add3}   & patche selection(local binary pattern), CNN      & Vision, Dresden     &    Image       \\\hline
MDM-CPS (2023) \cite{add4}   & Multi-distance measures  &Vision, Dresden       &    Image        \\\hline
Grad-CAM (2024) \cite{add6}   &  ResNet     & Vision     &    Video+ audio       \\\hline
\end{tabular}
\end{table*}
Multi-modal networks using different features have also been proposed to identify source cameras. For example, in \cite{add1}, the authors proposed a method for camera model identification (CMI) using a two-stream CNN model of RGB and noise images derived from high-pass filtering to enhance feature extraction. Akbari et al. \cite{add2} then presented a hierarchical Deep learning model to extract PRNU features. They designed a six-stream network to capture both low and high-level features, enhanced by a fusion layer. For verifying image integrity via camera model identification, the authors in \cite{add3} introduced a patch selection method using a uniform local binary pattern operator to enhance training data diversity and a dual-path enhanced ConvNeXt network that leverages multi-frequency information without a residual prediction module. The authors in \cite{add4} proposed a multi-distance measure and coordinate pseudo-label selection (MDM-CPS) method to tackle the challenge of limited training samples. Utilizing semi-supervised learning, MDM-CPS iteratively improves the labeled database while reducing the effects of noisy pseudo-labels. The authors in \cite{add6} proposed log-Mel spectrograms from audio, combined with an optimized ResNet model and Gradient-weighted Class Activation Mapping (Grad-CAM) to highlight key areas. By focusing on high-frequency components and applying band-pass filtering, the method achieves high accuracy in camera model classification.

In other works, researchers attempted to identify specific camera models using different deep learning methods. The difference between these approaches and the aforementioned ones is that these methods select some specific models from different datasets and then use them for training. Each method, however, addresses some specific challenges and try to solve them from a different perspective. For example, in \cite{add7} the authors used deep CNNs and machine learning algorithms to assess image authenticity and identify camera models on a dataset with established limits on image counts per model. While in \cite{add8} the authors employed multi-modal CNNs, integrating both audio and visual data to enhance source identification in video sequences. In the same context, a generalized framework using integral image optimization and constrained neural networks to mitigate semantic interference in feature extraction is proposed in \cite{add9}. To improve source camera identification accuracy by employing a multi-class ensemble classifier, the authors in \cite{add10} focused on exploiting demosaicing information extracted from images. In \cite{add11}, the authors introduced a classifier-block-level hierarchical system to reduce the complexity and memory usage while identifying camera brands based on intrinsic noise patterns. In \cite{add12}, the authors analyzed fingerprints of digital devices and images in IoT environments using CNNs for effective source identification, achieving high accuracy. The authors in \cite{add13} used Gaussian processes (GPs) for open-set camera model identification, providing built-in rejection mechanisms and reliable uncertainty estimates. To enhance CNN performance and in camera model identification, a selective pre-processing method is introduced in \cite{add14}. In \cite{add15}, the authors explored digital images for camera model identification. They criticized the reliance on a single database (Dresden) for deep learning evaluations. Their approach was to test multiple public databases (Dresden, SOCRatES, and Forchheim) and compare different deep learning methods with a particular focus on identification with transfer learning and fine-tuning for analysis.

In summary, the various methods proposed for camera model identification and source device identification offer significant advancements in digital forensics and multimedia analysis. Techniques such as dual-branch convolutional neural networks, hierarchical deep learning models, Discriminative Feature Projection, as well as the deployment of different CNN blocks have shown notable success in addressing challenges with limited training samples, and enhancing identification accuracy using images and videos. These methods effectively leverage deep learning, feature extraction, and semi-supervised learning strategies to boost performance, even under real-world conditions, such as social media processing. Collectively, they represent a substantial progression in the reliability, accuracy, and interpretability of source camera identification, ultimately offering solutions to law enforcement organisations and forensic analysts. Table \ref{table_summ} summarises the state-of-the-art methods mentioned earlier as well as the datasets reported in their experiments.

\section{Proposed system}
\begin{figure*}[t!]
    \centering
    \includegraphics[width=1\linewidth]{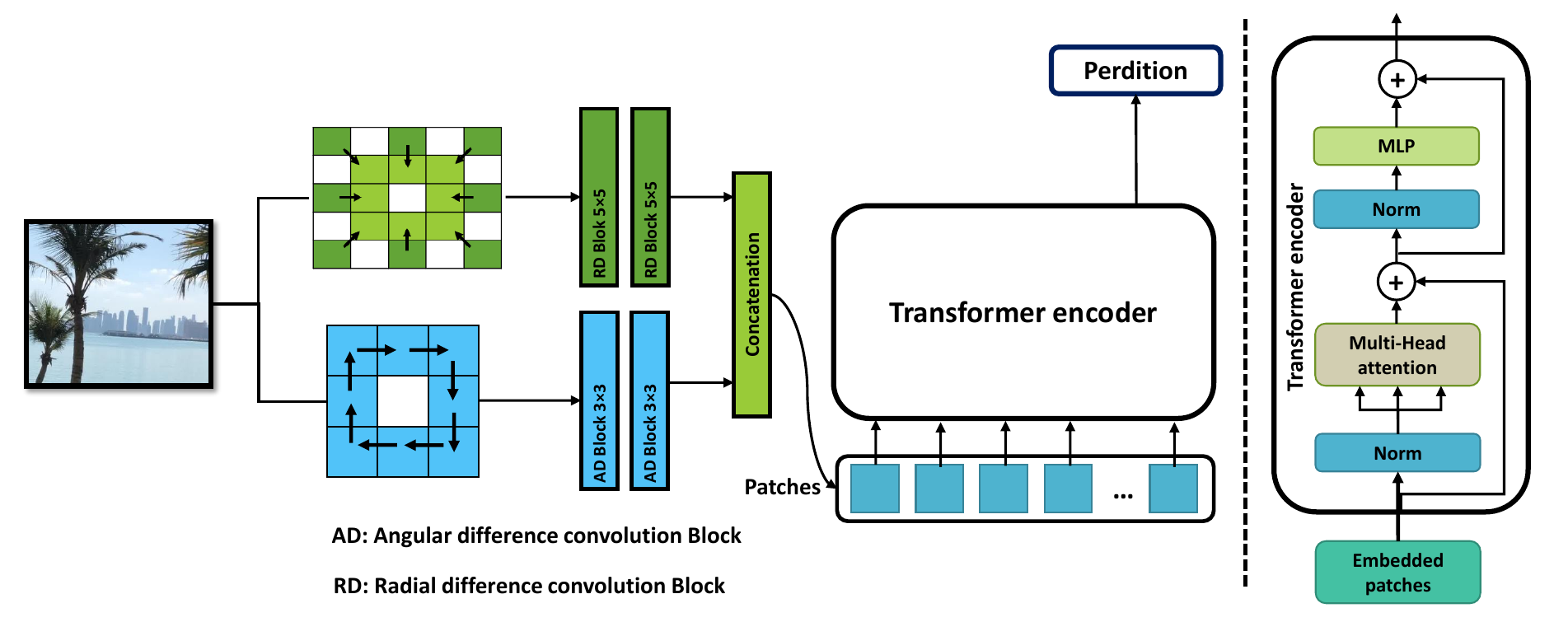}
    \begin{tabular}[b]{c}
        \includegraphics[angle=90,width=.41\linewidth]{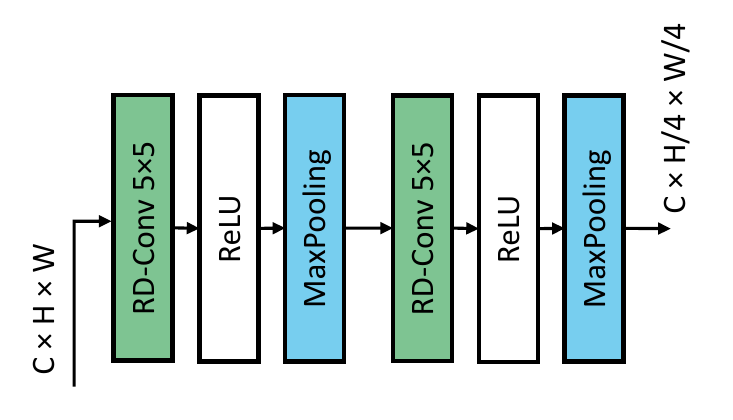}\\
        Angular Difference Block
    \end{tabular}
    \begin{tabular}[b]{c}
    \includegraphics[angle=90,width=.41\linewidth]{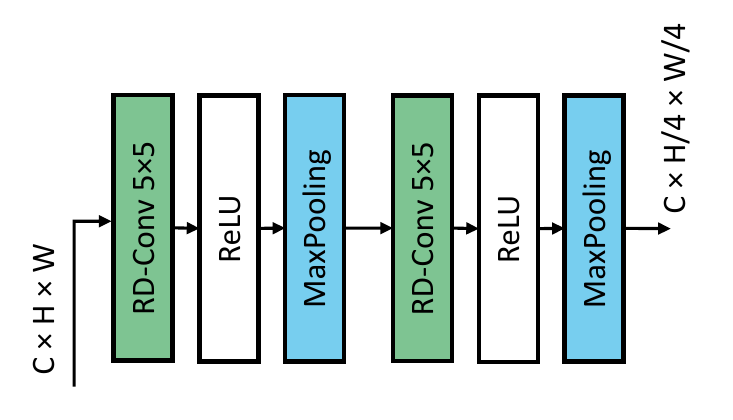}\\
      Radial Difference Block
    \end{tabular}
    \caption{Flowchart of the proposed network.}
   \label{fig_model}
\end{figure*}

This paper proposes a pixel-based deep learning method combining Pixel Difference convolution (PDC) for feature extraction and a Vision Transformer (ViT) network for classification. Unlike the existing methods, which are based on existing backbones like ResNet or MobileNet for extracting the features, the PDC features including Angular PDC (APDC) and Radial PDC (RPDC) are employed to efficiently extract the features of images taken using several devices of various make and models. Figure \ref{fig_model} represent each part of our system including the PDC feature extraction module and the classification module with Vision Transformer. The proposed PDC-ViT model is described in detail in the following sections.

\subsection{Pixel Difference Convolution Backbone}

The source camera identification from an image of the same scene can be difficult using content analysis, but the analysis of the pixel variations and the regional connections between the pixels (pixel neighboring analysis) can be an attractive alternative. In the literature, the processing of these images using filters and convolutional operations can generate this difference in terms of the distribution of pixels as well as the patterns of connections between them. For example, using the concept of kernels can make the identification task achievable with a good performance. From these operations, one can utilize several filters including radial, central or angular pixel difference \cite{pr1,pr2,pr3}. These techniques are introduced as kernels of convolutional layers. Pixel difference convolution (PDC) with vanilla convolution is one of the techniques that was used for face detection and recognition, edge detection, and many other computer vision tasks.
We aim to calculate the similarity between PDC and the vanilla convolution process by considering the kernels used in vanilla convolutions by using the difference operation between neighboring pixels using angular, radial also central ways. The vanilla convolution can be expressed as follows:

\begin{equation} \label{eq1}
   Y=  \sum_{n=1}^{9} w_n \cdot x_n
\end{equation}

\begin{figure}[t!]
    \centering
    \includegraphics[width=.71\linewidth]{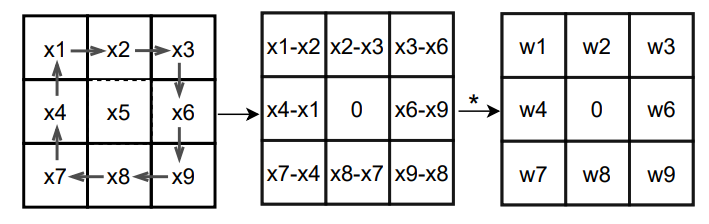}
    \caption{Selection of pixel pairs and convolution in Angular PDC.}
   \label{apdc}
\end{figure}

Where $w_n$ and $x_n$ denote the weight value in the convolution kernel and input pixels, respectively. $Y$ denotes the central pixel location of local patches. The combination of the features from a local (region-based) variation using the difference between neighboring pixels, and holistic attributes can be useful for detecting patterns resulting in an improved learning process. To achieve this, the use of pixel difference convolution exploding the angular and radial differences can augment the distribution of local patterns and select the vanilla convolution gradients using 3 × 3 kernels for Angular PDC (APDC) and  5×5 kernels for Radial PDC (APDC). These kernels are used to propagate local patches for a smoothing of the local detailed cues \cite{pr1}. As reported in \cite{pr2}, and using the same approach, the subtracted pixels in the window are generated as used in vanilla convolution. The PDC equation, in a set of pixel pairs picked from the current local patch \{($x_1$,$x_1^{'}$),($x_2$,$x_2^{'}$),...,($x_m$,$x_m^{'}$) \}, is based on vanilla convolution in Equation \ref{eq1} , can be expressed as follows  :

\begin{equation} \label{eq4}
   Y=  \sum_{(x_i,x_i^{'}) \in \mathcal{P}} w_n \cdot (x_n-x_n^{'})
\end{equation}

 where $x_n$ and $x_n^{'}$ denotes the input pixels, $w_n$ denotes the k × k convolution kernel weight. Then, the gradient information can be captured by selecting the pixel pairs using angular or radial operations. This can lead to high-level feature extraction. Unlike the method that uses of static filters, PDC can learn from different scales during the training process that different gradient information will be detected from images. Taking APDC as an example illustrated in Figure \ref{apdc}, the conversion is expressed by the following equations:

\begin{equation}
\begin{split}
 y =& w_1\cdot (x_1-x_2)+w_2\cdot (x_2-x_3)+w_3\cdot (x_3-x_6)+...+ w_4\cdot (x_4-x_1)\\
   =&(w_1-w_4)\cdot x_1+(w_2-w_1)\cdot x_2+(w_3-w_6)\cdot x_3+...\\ =&\hat{w_1} \cdot x_1+\hat{w_2} \cdot x_2+\hat{w_3} \cdot x_3+...=\sum \hat{w_i} \cdot x_i
\end{split}
\end{equation}

Using This PDC representation, a network starts by computing the differences of kernel weights, followed by some simple convolutions to extract more learning features if the input features maps are unchanged. 

\subsection{Proposed PDC-ViT  Model}

With the introduction of attention modules and their attractive performances for Neural Language Processing (NLP) tasks, computer vision researchers have attempted to combine self-attention with CNN-based architectures to further improve the performance of image processing tasks \cite{pr4}. This led to the wide use of attention instead of convolutions networks for recognition or classification purposes. In addition, it has become the basis of creating another technique based on attentions named transformer which worked well in NLP \cite{pr4,pr5}. Due to the successes of the transformer networks in Neural Language Processing (NLP), researchers have tried to exploit this technique concept for image recognition, image classification, and object detection tasks \cite{tr1}. To be suitable for images, transformer networks used in NLP have been modified and adapted where images are divided into patches to be embedded linearly as input of the transformer network \cite{tr2}. These image patches are represented as the tokens (in NLP) where the transformer process begins by dividing the images into patches, then flattening followed by a linear embedding of the patches \cite{tr3}. The generated sequence then is introduced into the transformer encoder for clarification using supervised learning.

Vision Transformers (ViT) are one of the popular network types that use transformers and have gained significant attention in the field of image processing and computer vision, especially in image classification tasks. One of the main advantages of Vision Transformers is their ability to capture global context information and long-range dependencies in images, which allows a better image understanding. Also, the ViT networks can perform well on diverse datasets and adapt to different domains without significant modifications. ViT networks achieved competitive results on various image classification benchmarks, even surpassing some convolutional neural network (CNN) architectures. 

In this paper, we propose a source camera identification method using PDC features and Vision Transformer. Rather than using the patches of the original images as input of the Vision Transformer network, we exploit PDC as the main feature extraction using the angular and radial features as described previously. The extracted features are then used as input of the Vision Transformer module used in \cite{tr1} with a multi-head attention network where the patch size is set to $64\times64$. This combination allows for a better learning process. In addition, this results in a training process with a low number of parameters as well as specific PDC features for source camera identification compared to ordinary features provided by ResNet or VGG. 

The scientific novelty of our work primarily lies in the introduction of the Pixel Difference Convolution (PDC) method for feature extraction, which provides a unique pixel-based analysis in contrast to existing approaches that rely predominantly on conventional deep learning frameworks. By using PDC, our methodology enhances the ability to capture critical pixel variations and neighbourhood relationships, thus significantly improving the accuracy of source camera identification.

The integration of the Vision Transformer for classification further distinguishes our approach, as it offers a sophisticated way to process and analyze the extracted features, making it more efficient in handling diverse datasets with various brands and models. This innovative combination not only addresses the current challenges in source camera identification but also gives an introduction for future engineering applications within AI.

Moreover, our method can be adapted for real-time forensic analysis. As camera technology continues to evolve, our system can be further refined and extended to keep pace with new developments, offering substantial benefits in a wide range of disciplines such as cyber security, forensic science, and AI. Overall, the novel aspects of this work provide a strong foundation for future insights in AI applications.

\subsection{Model Advantages and applications }

Integrating Pixel Difference Convolution (PDC) features and Vision Transformers (ViT) for source camera identification offers several significant advantages. Because it analyzes pixel variations and regional correlations, which are key to the discriminative  power, PDC can be viewed as an effective feature extraction module. Also, PDC features can highlight subtle differences in pixel distributions and patterns that are often characteristics of specific cameras. While conventional content analysis methods struggle with images of the same scene taken by different devices, PDC can effectively capture and emphasize unique pixel-level variations introduced by distinct camera sensors. This can lead to improved performance in source camera identification tasks by allowing the model to utilize both global pattern recognition and local pixel variation analysis efficiently. Furthermore, the use of Vision Transformer as a classification technique allows the method to capture global context information and long-range dependencies, and hence perform accurately. Finally, the combination of PDC features and ViT can reduce the number of parameters needed during the training phase, making it computationally efficient without compromising the performance.

The use of Vision Transformers (ViT) with PDC features for source camera identification has some limitations, though. It requires large amounts of training data, which may not always be available, and can be computationally intensive, necessitating powerful hardware resources. Additionally, the model may struggle with low-quality images and could be prone to  overfitting if trained on limited data. Interpretability is also a concern, as the decision-making process of transformers can be less transparent.

In terms of applications, this work can benefit areas in source camera identification and media authentication, such as digital forensics, copyright management, and forgery detection. In digital forensics, the ability to trace the origin of an image can provide critical evidence in legal investigations. Also, the method can be used in authenticating digital media, helping to protect intellectual property rights and detect malicious forgeries. Overall, the fusion of Vision Transformers and PDC features opens new avenues for robust image analysis and forensics.

\section{Experimental results}
This section discusses the experimental results obtained by the proposed method using four public datasets including Vision, QUFVD, Daxing, SOCRATES, and Video-ACID. The evaluations and comparisons have been carried out to demonstrate the effectiveness of the proposed architecture against the existing methods.
The obtained results have been compared with those reported for a number of state-of-the-art techniques including \cite{r4}, \cite{r5}, \cite{r6}, \cite{r7}, \cite{r8}, \cite{i3}, \cite{i4}, \cite{i5}, and \cite{i6}. The results are also presented by visualizing the obtained confusion matrices, and the T-SNE visualization for each dataset.

\subsection{Implementation setup}


In order to train and evaluate the proposed method, the split protocol (train, validation, test) of each dataset has been respected to ensure a systematic approach to model training and performance assessment. While, we used 80\% of each dataset for Training/validation and 20\% for testing. The datasets contain frames taken from videos for use to train our proposed model. Also for all datasets, we pre-processed the images by cropping them to a standardized size of (224 × 224) pixels. This standardization facilitates uniformity across the learning phases, as well as optimizes the feature extraction process by providing a consistent input shape for the model.

The proposed system has been trained on a high-performance laptop equipped with 64 GB of RAM and an NVIDIA GeForce RTX 2070 graphics card, which allows handling the computational load associated with deep learning tasks. The implementation of the system was done using Python with PyTorch library.

For optimizing the model's performance, the CrossEntropy loss function was employed, which is particularly effective for multi-class classification tasks typically encountered in source camera identification scenarios. Additionally, the Adam optimizer was used, known for its adaptive learning rate capabilities, enhancing convergence speed and stability during training. The learning rate was set to a value of 0.00003, balancing the need for effective learning while minimizing the risk of overshooting the optimal solution.

For the PDC-ViT network, We used two blocks for each features including Angular PDC (APDC) block (with 3 × 3 kernels) and Radial PDC (APDC) block (with 5×5 kernels). For the Vision Transformer network parameters, we set dim to 1024, depth to 6, heads to 16, mlp\_dim to 2048, dropout to 0.1, and emb\_dropout to 0.1.

\subsection{Datasets}

Vision dataset introduced in \cite{vision}, which is the most popular database for source camera identification, contains 35 smartphone devices from 11 major brands with 34,427 images and 1914 videos. The video captured different scenes including indoor, outdoor, and flat scenarios. Some videos are taken in still mode while the user stands still while the video is recorded. The other videos represent scenes when the user walking. 
\begin{table}[!t]
\renewcommand{\arraystretch}{1.3}
\caption{Performance of source camera identification methods on Vision dataset. The \textbf{bold} and \underline{underline} fonts respectively represent the \textbf{first} and \underline{second} place}
\label{table_vision}
\centering
\begin{tabular}{l|c|c|c}
\hline 
\textbf{Method} &  \textbf{Resolution}& \textbf{\# Devices}&\textbf{Accuracy}\\
\hline
Shullani et al. (2017) \cite{vision}& 960 × 720& 35& 83.00\\
\hline
Xiao et al. (2022) \cite{r4}&  512 × 512& 35&81.10\\
\cline{2-4}
& 256 × 256& 35&66.20 \\
\hline
Shallow CNN (2018) \cite{r6} & 64 × 64 ×3& 35&32.87\\
\hline
DenseNet-40 (2018) \cite{r6} & 32 × 32 ×3 &35& 76.78\\
\hline
DenseNet-12 (2018) \cite{r6} & 224 × 224 ×3 & 35&85.06\\
\hline
XceptionNet (2018) \cite{r6} &299 × 299 ×3 &35& \underline{91.36}\\
\hline
Sameer et al. (2022)\cite{r7} &Varied  & 11& 75.20 \\
\hline
ResNet50 (3×3) (2022) \cite{r8}& 480 × 800 ×3& 28&67.81  \\ \cline{3-4}
ResNet50 (5×5) (2022) \cite{r8} &&28&54.18  \\\cline{3-4}
MobileNet (3×3) (2022) \cite{r8}&&28& 72.47\\\cline{3-4}
MobileNet (5×5) (2022) \cite{r8}&&28& 53.19  \\\hline

Huan et al. (2024) \cite{add3} &64 × 64 ×3& 28& 81.30 \\ \hline
MDM-CPS (2023) \cite{add4}&-& 35& 87.74 \\ \hline

\textbf{PDC-ViT  (Ours)} &224×224×3&35& \textbf{94.30} \\
\hline
\end{tabular}
\end{table} 

\begin{figure*}[t!]
    \centering
    \begin{subfigure}[b]{0.48\textwidth}
        \includegraphics[width=\textwidth]{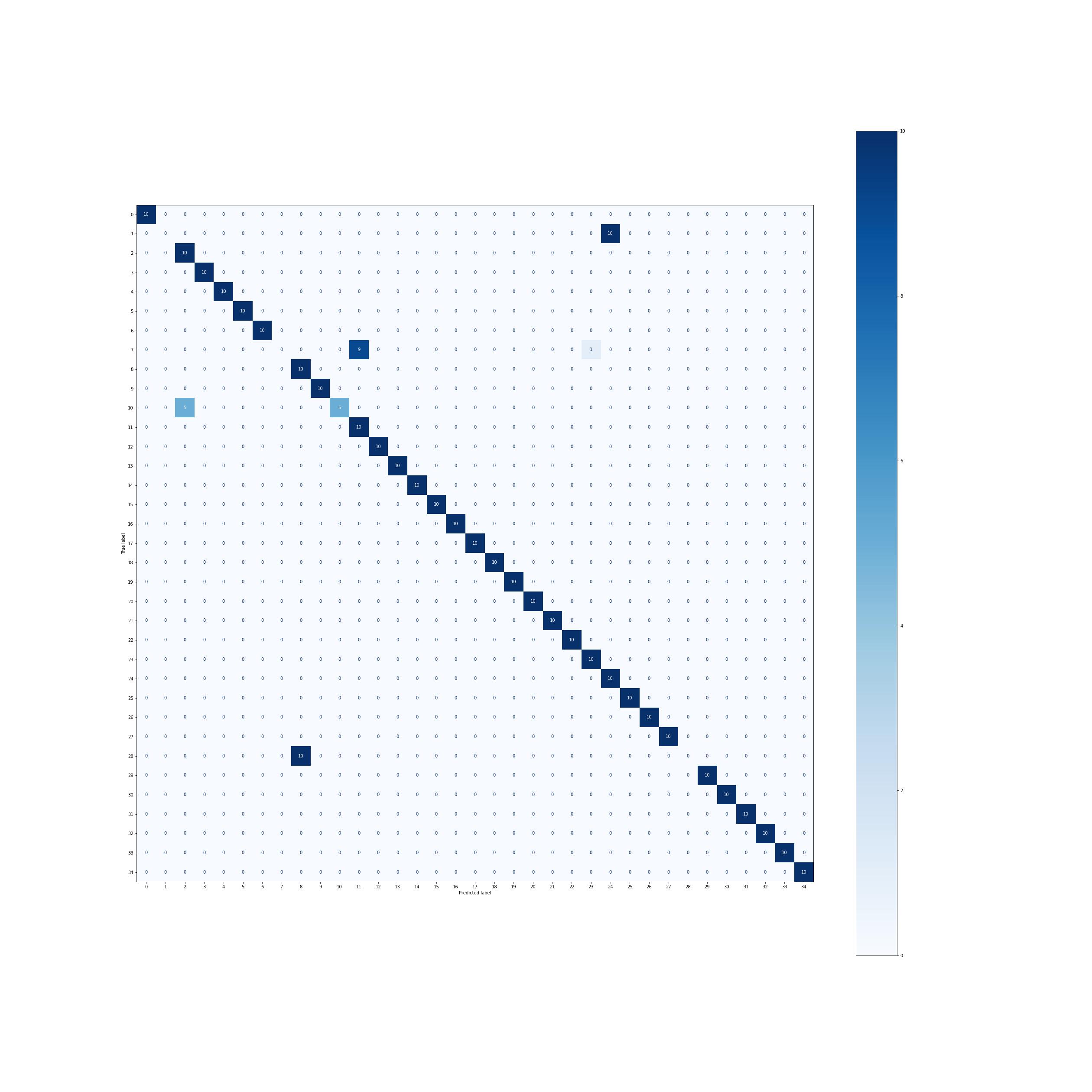}
        \caption{Vision dataset}
    \end{subfigure}
    \begin{subfigure}[b]{0.48\textwidth}
        \includegraphics[width=\textwidth]{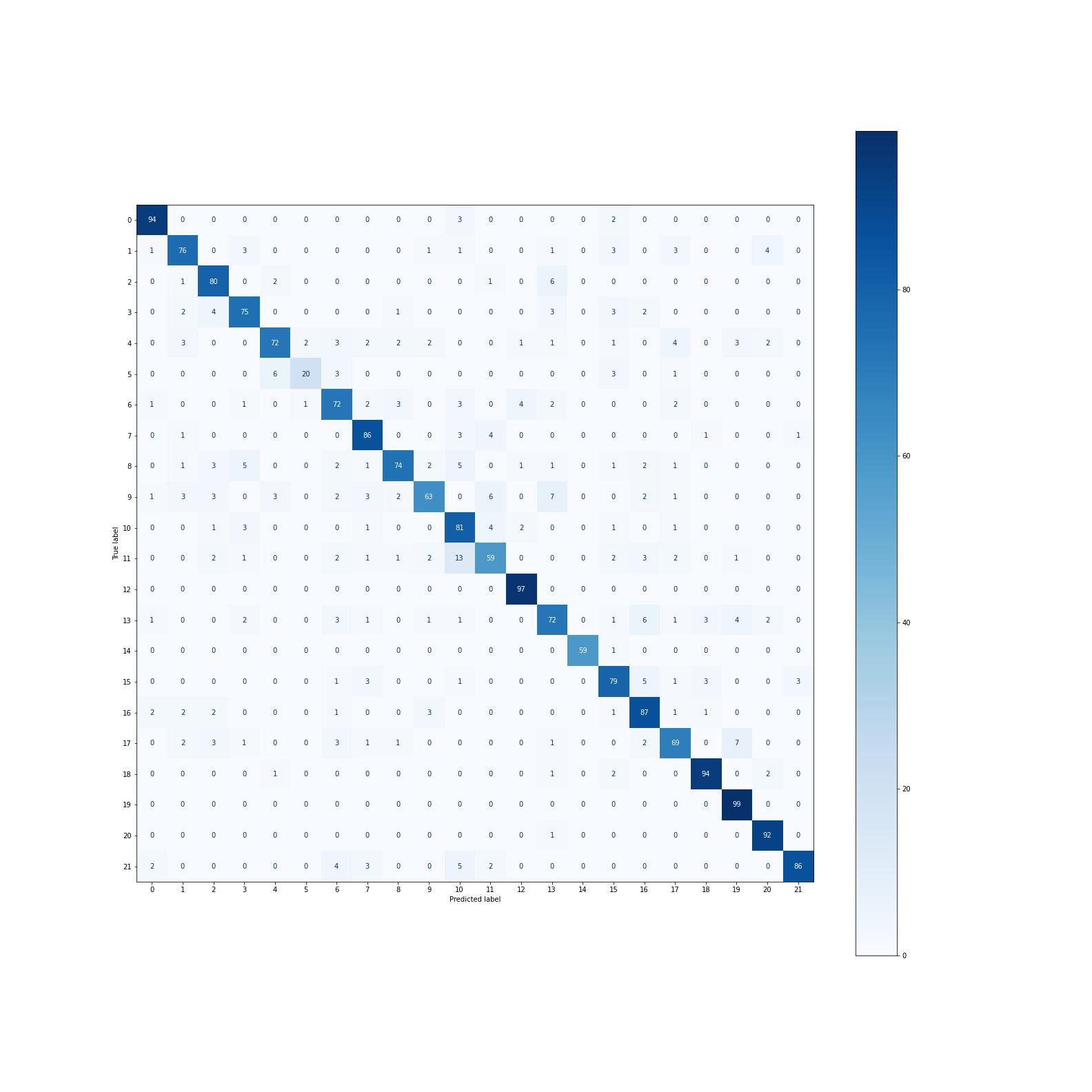}
        \caption{Daxing dataset}
    \end{subfigure}  

    \caption{Confusion matrices obtained of the evaluation using PDC-ViT on Vision and Daxing datasets. }
    \label{vision-daxing}
\end{figure*}
The Video-ACID dataset \cite{acid} is a public source camera identification dataset. Over 12,000 videos were collected from 46 physical cameras representing 36 different camera models in the video-ACID database. All of these videos were shot manually to represent a range of lighting conditions, content, and motion.

The Daxing smartphone identification database \cite{daxing} includes both images and videos captured from various smartphones of different brands and models. The data from 90 smartphones, representing 22 models and 5 brands, includes 43400 images and 1400 videos. In the case of the iPhone 6S (Plus), 23 different smartphone models are available. Scenes selected normally include a sky, grass, rocks, trees, stairs, a vertical printer, a lobby wall, and a white wall in a classroom, among others. The videos were shot vertically in each scene where each scene contains at least three videos. In addition, all videos were recorded over 10 seconds. 

The SOCRatES dataset\cite{socrates} has been developed using smartphones. Approximately 9700 images and 1000 videos were taken by 103 different smartphones from 15 different brands.

\begin{table}[!t]
\renewcommand{\arraystretch}{1.3} 
\caption{Performance of source camera identification methods on Daxing dataset. The \textbf{bold} and \underline{underline} fonts respectively represent the \textbf{first} and \underline{second} place.}
\label{table_daxing}
\centering
\begin{tabular}{l|p{2.5cm}|p{2.5cm}}
\hline
\textbf{Method}& \textbf{Resolution} & \textbf{Accuracy}\\
\hline 

xDnCNN (2022) \cite{r44}& 128×128 &24.21\\  \cline{2-3}
               &256×256  & 48.26\\ \cline{2-3}
               &512×512  & 68.90\\ \hline 
DHDN (2022) \cite{r4}& 128×128 & 46.20\\  \cline{2-3}
               &256×256  & 66.21\\ \cline{2-3}
               &512×512  & \underline{81.11}\\ \hline 
Akbari et al. (2024) \cite{add2}& 350×350 &69.9\\  \hline
              
\textbf{PDC-ViT  (Ours)} &224×224& \textbf{84.06} \\ \hline

\end{tabular}
\end{table}

QUFVD dataset \cite{QUFVD} includes 6000 videos from 20 modern smartphones representing five brands where each having two models, and each model has two identical smartphone devices. This database is suitable for
deep learning methods, and the results from this database show that new databases with devices based on new technologies need more improvement in both ISCI and SCMI scenarios.
\begin{figure*}[t!]
    \centering
    \begin{subfigure}[b]{0.43\textwidth}
        \includegraphics[width=\textwidth]{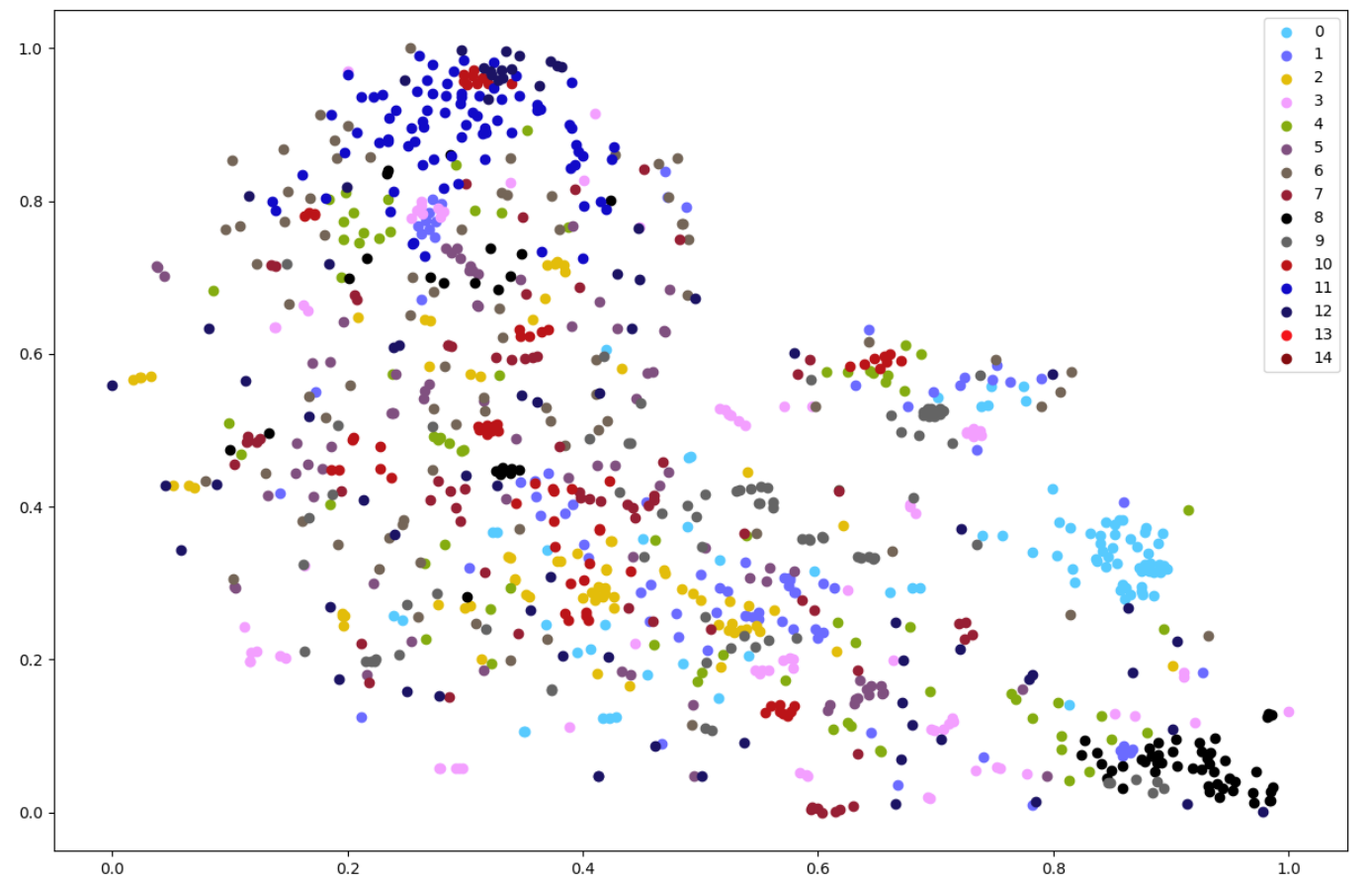}
        \caption{t-SNE of ResNet101 on Socrates.}
    \end{subfigure}
    \begin{subfigure}[b]{0.43\textwidth}
        \includegraphics[width=\textwidth]{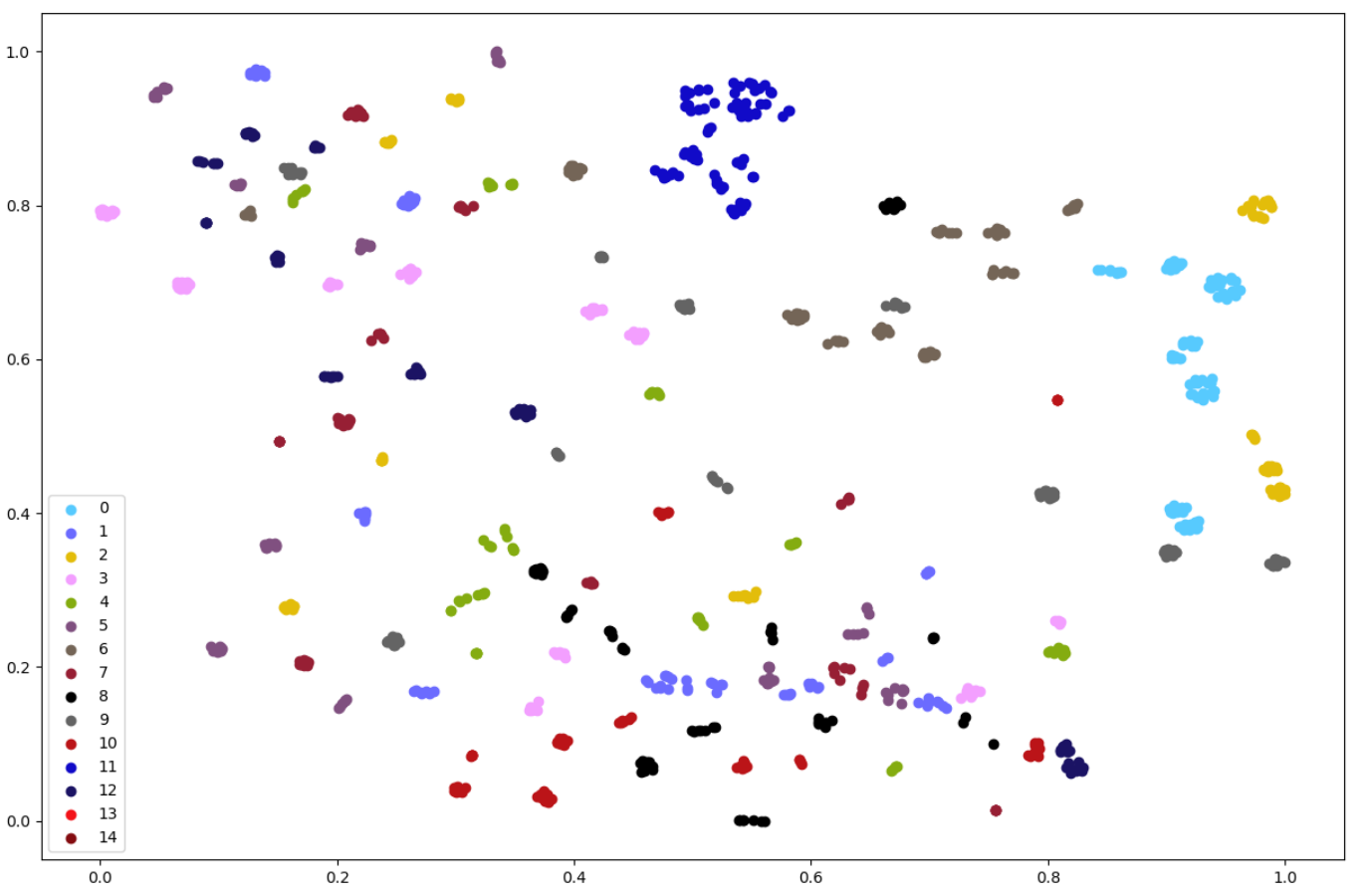}
        \caption{t-SNE PDC-ViT  on Socrates}
    \end{subfigure}  
    \begin{subfigure}[b]{0.43\textwidth}
        \includegraphics[width=\textwidth]{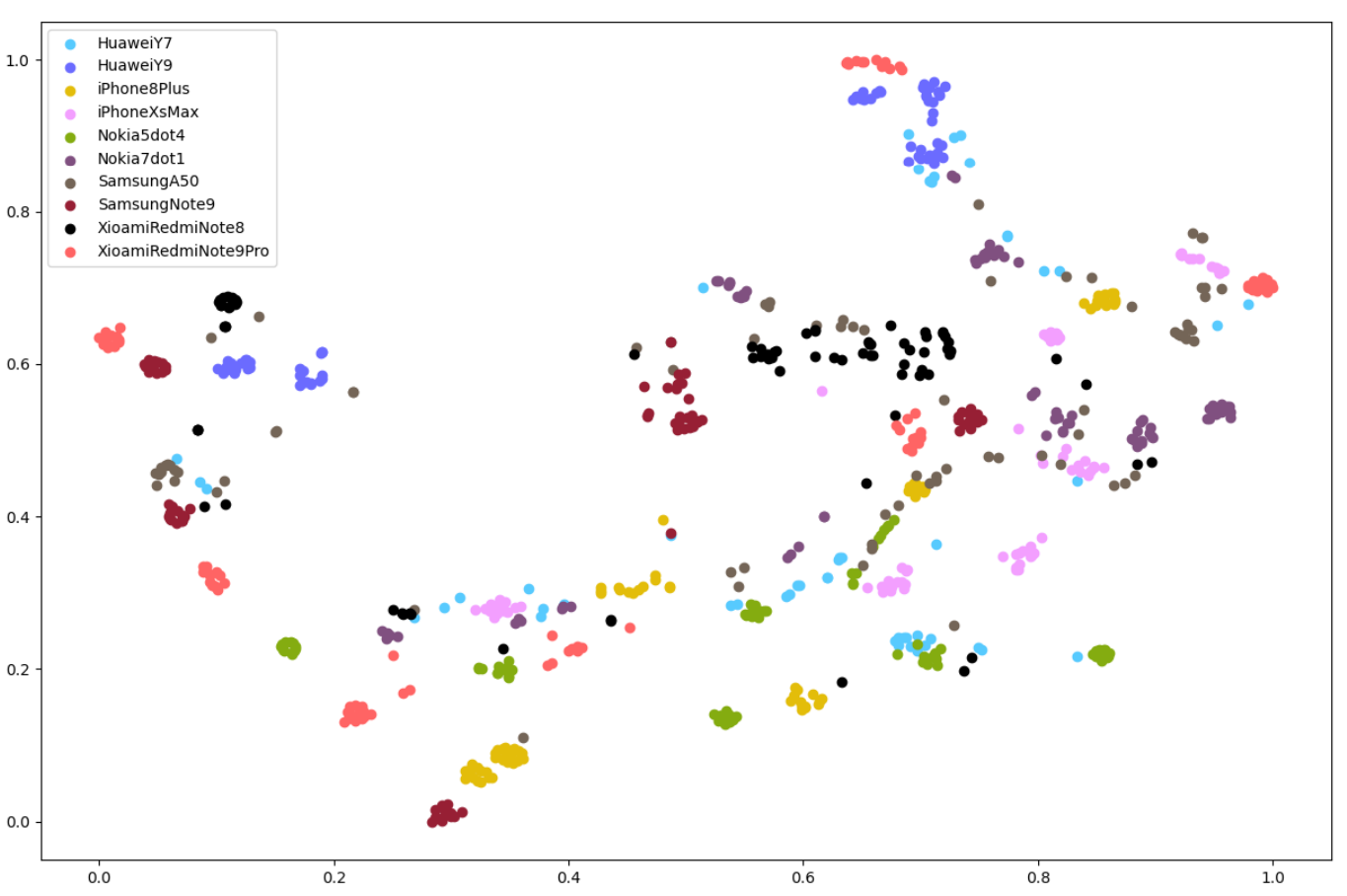}
        \caption{t-SNE of ResNet101 on QUFVD.}
    \end{subfigure}
    \begin{subfigure}[b]{0.43\textwidth}
        \includegraphics[width=\textwidth]{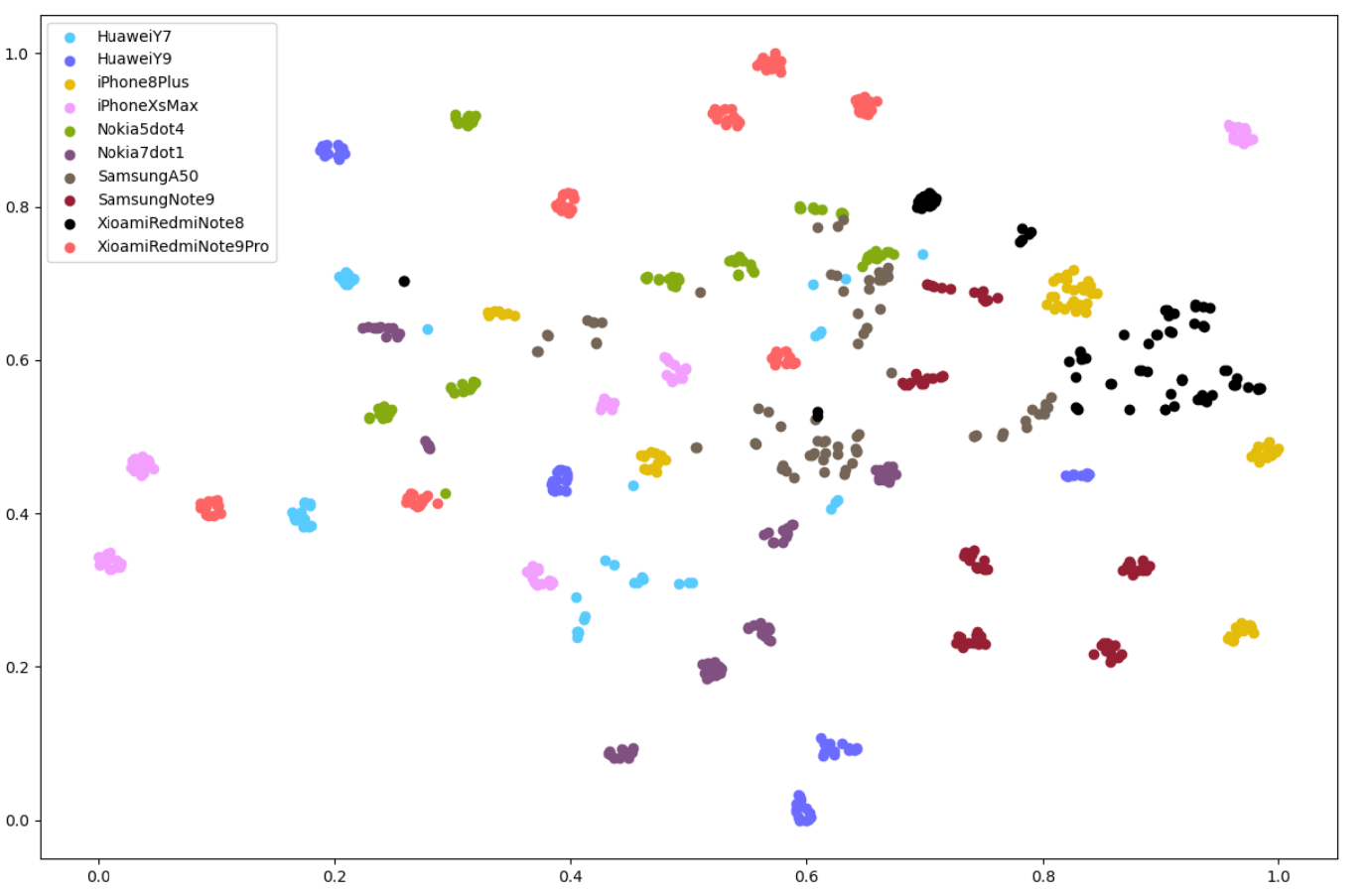}
        \caption{t-SNE of PDC-ViT on QUFVD.}
    \end{subfigure}
    \caption{t-SNE visualization on QUFVD and Vision dataset }
    \label{tsne}
\end{figure*}

\subsection{Evaluation using the Vision dataset}
To evaluate the proposed source camera identification method,  35000  images of 35 different devices have been used for training and testing. The images used are first cropped with a size of $224\times 224$ before starting the training process. For the testing phase, the images are selected randomly from each set of cameras. The obtained results using the state-of-the-art methods as well as the proposed method are shown in Table \ref{table_vision}. The Tables represent the number of devices used for evaluation, the image sizes used for training, and the performance accuracy for different approaches. From The table, it can be observed that the proposed (PDC-ViT ) method reached the best accuracy results compared with the other approaches with an accuracy of 94.30\%, outperforming the method in \cite{r6} by 3.24\%  which used the XceptionNet network. In terms of image size used for training, PDC-ViT and DenseNet-1 \cite{r6} used the same image size $224\times 224$ and the performance accuracy reached 94.30\% and 85.06\%, respectively. This demonstrates that PDC-ViT outperforms the other methods even with the same image size. The same observation holds if we compare PDC-ViT on the method that uses a smaller number of devices in the evaluation like in \cite{r7}, ResNet50 (3×3) \cite{r8}, ResNet50 (5×5) \cite{r8} , MobileNet (3×3) \cite{r8}, and MobileNet (5×5) \cite{r8}. Also, the proposed method outperforms the method described in \cite{r7} by 19.1\% and with much higher performances for the other methods. Furthermore, it can be suggested from these experiments that the PDC techniques combined with the Vision Transformer network can identify the source cameras with high-performance accuracy even with the high number of camera models. Also, the use of pixel-based processing provides a much more attractive tool to differentiate between camera devices. Also, the numerical results are demonstrated using the confusion matrix as illustrated in Figure \ref{vision-daxing} (a) which represents the number of positive and negative camera models identification.
\begin{table}[t!]
\renewcommand{\arraystretch}{1.3}
\caption{Performance of source camera identification methods on Socrates dataset. The \textbf{bold} and \underline{underline} fonts respectively represent the \textbf{first} and \underline{second} place}
\label{table_soc}
\centering

\begin{tabular}{p{4cm}|p{2.5cm}}
\hline
\textbf{Method} & \textbf{Accuracy}\\
\hline 
Galdi et al. (2019) \cite{socrates}& 86.00\\ \hline 
Bayar et al. (2017) \cite{i3}& 75.20\\ \hline 
Bayar-MFR (2018) \cite{i4}& 75.05\\  \hline  
Mayer et al. (2020) \cite{i5}& \underline{92.31}\\ \hline 
Ding et al. (2019) \cite{i6}& 68.08\\ \hline 
\textbf{PDC-ViT  (Ours)} & \textbf{94.22} \\ \hline 
\end{tabular}
\end{table}

\begin{table}[!t]
\renewcommand{\arraystretch}{1.3}
\caption{Performance of source camera identification methods on QUFVD dataset. The \textbf{bold} and \underline{underline} fonts respectively represent the \textbf{first} and \underline{second} place}
\label{table_QUFVD}
\centering
\begin{tabular}{l|c|c}
\hline
\textbf{Method} &\textbf{Resolution} &\textbf{Accuracy}\\
\hline 
MISLNet (2022) \cite{QUFVD} & 350 × 350 ×1&59.60\\
\hline 
MISLNet  (2022) \cite{QUFVD}& 350 × 350 ×3&51.20\\\hline 
MobileNet (2022) \cite{r8} &480 × 800 ×3 &71.75\\\hline 
MobileNet (5x5) (2022) \cite{r8}& 480 × 800 ×3&70.50\\\hline
 
Akbari et al. (2024) \cite{add2}& 350×350×3&  81.6 \\\hline
\textbf{RPDC-ViT (Ours)} &224×224×3 &83.36\\\hline
\textbf{APDC-ViT (Ours)} &224×224×3 &\underline{85.47}\\\hline
\textbf{PDC-ViT  (Ours)} & 224×224×3&\textbf{92.29}\\\hline

\end{tabular}
\end{table}

\begin{figure*}[t!]
    \centering
    \begin{subfigure}[b]{0.32\textwidth}
        \includegraphics[width=\textwidth]{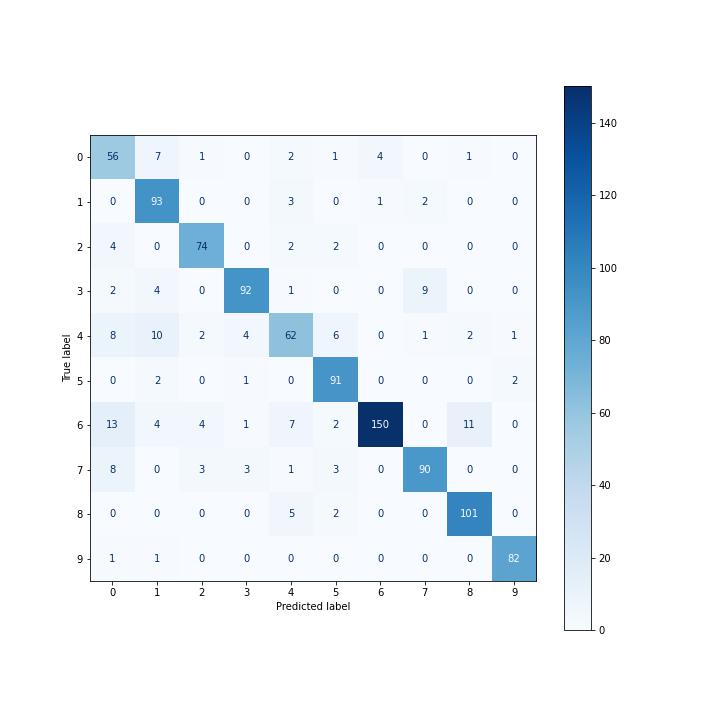}
        \caption{APDC-ViT }
    \end{subfigure}
    \begin{subfigure}[b]{0.32\textwidth}
        \includegraphics[width=\textwidth]{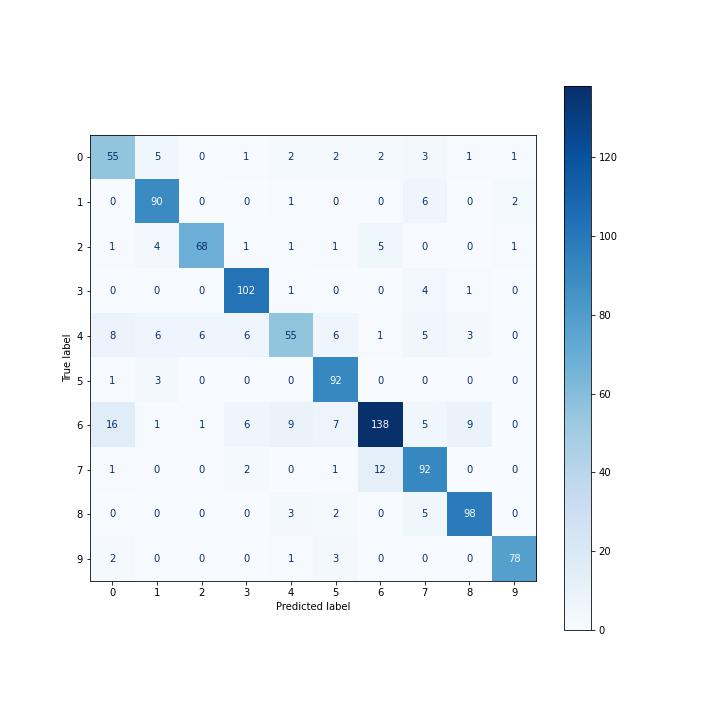}
        \caption{RPDC-ViT }
    \end{subfigure}  
    \begin{subfigure}[b]{0.32\textwidth}
        \includegraphics[width=\textwidth]{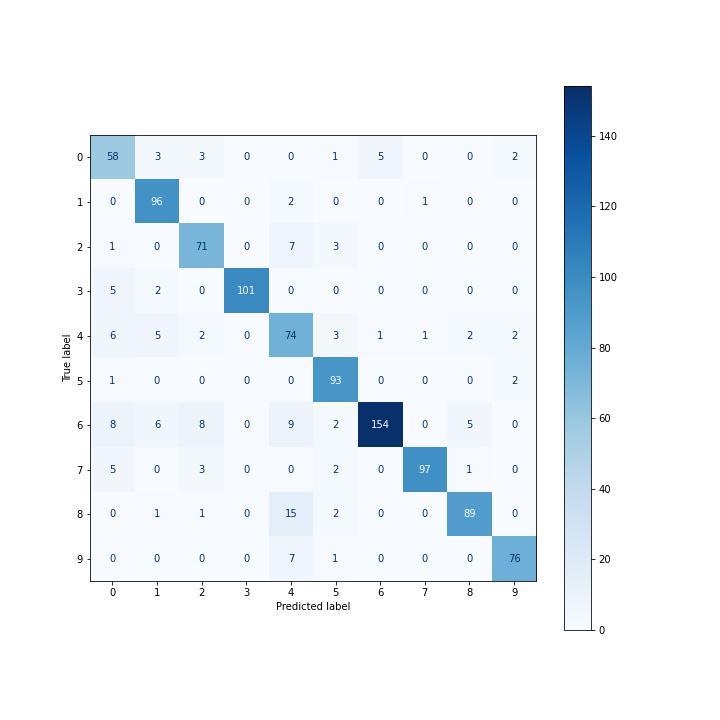}
        \caption{PDC-ViT  (ARPDC-ViT )}
    \end{subfigure}
    \caption{Confusion matrices obtained of the evaluation using APDC-ViT, RPDC-ViT, and PDC-ViT  (ARPDC-ViT ) proposed architectures on QUFVD dataset. }
    \label{confQUFVD}
\end{figure*}

\subsection{Evaluation using the Daxing dataset}
To evaluate the source camera identification with the numerical values, both the performance accuracy and the confusion matrix methods have been used. Table \ref{table_daxing} and Figure \ref{vision-daxing} (b) depict the evaluation of our proposed method and its comparison against some state-of-the-art methods. Table \ref{table_daxing} demonstrates that the proposed PDC-ViT architecture provides a better performance accuracy than xDnCNN \cite{r44} and DHDN \cite{r44} for different image resolutions. The performance accuracy of the PDC-ViT method reached 82.81 using images of size $224\times 224$ with a performance improvement of 2.95\% compared to the second best value achieved using DHDN whit an image size of $512\times 512$. In addition, it outperforms the method using xDnCNN \cite{r44} with the same image size. From the Table, it can be seen that the image size used in training the proposed models has an impact on the obtained performance accuracy. However, if we compare our obtained accuracy with an image size of $224\times 224$ is further improved compared with larger image sizes of $256\times 256$ and $512\times 512$. Also, it is worth noting that a performance accuracy improvement of about 39\% was achieved with images of size $128\times 128$ for the proposed architecture. The results in Table \ref{table_daxing} depict the performances in terms of the confusion matrix showing a successful identification of the source camera with minor errors.

\subsection{Evaluation using the Socrates dataset}

Using the same evaluation protocol as above, experiments on the Socrates dataset including a comparative study with five state-of-the-art counterparts described in \cite{socrates}, \cite{i3}, \cite{i4}, \cite{i5}, and \cite{i6} have been carried out. Our proposed method has been trained and tested on the 60 camera models of the Socrates dataset. Table \ref{table_soc} shows the performance accuracy of each method including the proposed PDC-ViT. For the Table, From the Table, it can be seen that PDCViT reached an accuracy of 94.22\% outperforming its competitors. For example, PDC-ViT is better than the second-best accuracy reached using  \cite{i5} by a value of 2.9\%, while the difference is up to 8.2\% for the others.

\begin{figure}[t!]
    \centering
    \begin{subfigure}[b]{.51\textwidth}
        \includegraphics[width=\textwidth]{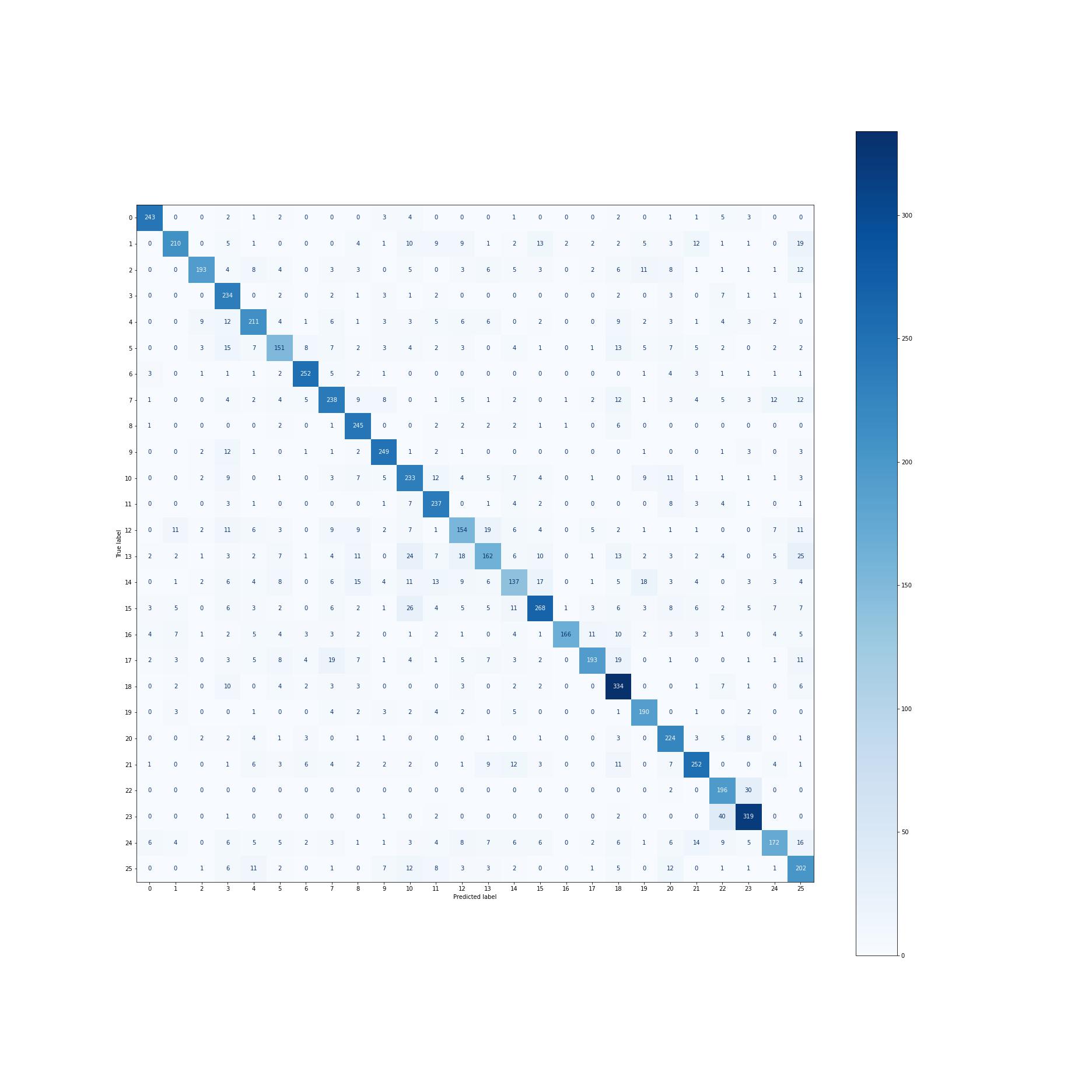}
        
    \end{subfigure}
    \caption{Confusion matrix using PDC-ViT  on Video-ACID dataset}
    \label{conf-acid}
\end{figure}

The PDC features are also compared with widely used classification networks such as ResNet101 by presenting t-SNE visualization in Figure \ref{tsne}. For Figure \ref{tsne} (a) and (b), it can be seen that the PDC features are more categorized that the ResNet101 counterparts. This clearly supports the results shown in Table \ref{table_soc} where the performance accuracy using the ResNet network is smaller than our proposed method. 

\subsection{Evaluation using the QUFVD dataset}
For QUFVD dataset, the evaluation of the source camera identification is carried out using performance accuracy including T-SNE visualization of ARPDC features. To achieve this, the proposed method using PDC-ViT  (Angular and Radial, APDC-ViT using Angular Pixel Difference only, and RPDC-ViT using Radial Pixel Difference were carried out. The obtained results are compared with the results of the state-of-the-art methods which are presented in Table \ref{table_QUFVD}.
The results clearly demonstrate that the proposed architecture using the three scenarios above outperforms the other methods achieving an accuracy reaching  92.29\%, 85.47\%, and 83.36\% for PDC-ViT, RPDC-ViT, and APDC-ViT,  respectively. It can also be noticed that the other methods used a patch size up to $224\times 224$, even though the performance accuracy using the proposed method is much better than the works described in \cite{QUFVD} which uses a size of $350\times 350$ with RGB and gray-scale representation and in \cite{r8} which uses a patch of size $480\times 800$ with MobileNet and Constraint MobileNet networks. The obtained results using numerical values are also demonstrated using confusion matrices in Figure \ref{confQUFVD} and t-SNE visualization protocol as described in Figure \ref{tsne} (d) which illustrates the distribution of the camera model using PDC features compared with the features of the same camera models but using ResNet101. This visualization result illustrates the ability of the proposed model to cluster the camera models in classes in an effective way compared to ResNet101 features visualization of the same data.

\begin{table}[t!]
\footnotesize
\centering
\renewcommand{\arraystretch}{1}
\caption{False Negative Rate (FNR) and False Positive Rate (FPR) of the proposed PDC-ViT and the state-of-the-art methods on 18 devices of Video-ACID dataset and 8 devices added in \cite{r10}. The \textbf{bold} and \underline{underline} fonts respectively represent the \textbf{first} and \underline{second} place.}
\label{table_acid}
  \begin{tabular}{p{1cm}|c|c|c|c|c|c|c}
    \hline
 \multirow{2}{*}{\textbf{Dataset }}  &  \multirow{2}{*}{\textbf{Device}} &
      \multicolumn{3}{c}{\textbf{FNR}} &
      \multicolumn{3}{c}{\textbf{FPR}} 
       \\ \cline{3-8}
       
&& \textbf{PRNU \cite{r10} }& \textbf{WA-PRNU \cite{r10}} & \textbf{PDC-ViT }&  \textbf{PRNU \cite{r10}} & \textbf{WA-PRNU \cite{r10}} & \textbf{PDC-ViT }\\ \hline
                                             
\multirow{18}{*}{\rotatebox[origin=c]{90}{\textbf{Video-ACID}} }&M1 & 9.8  & 12.7 &  9.3 &5.1  & 1.6 &  0.3   \\\cline{2-8}
&M2 & 100  & 91.8 &  32.6 & 0.1 & 1.9 &  0.5   \\\cline{2-8}
&M3 & 98.5 & 94.2 &  31.1 &0.2  & 0.4 &  0.3  \\\cline{2-8}
&M4 & 2.1  & 2.1  &  10.0 &8.8  & 6.1 &  1.6   \\\cline{2-8}
&M5 & 40.6 & 31.9 &  27.9 &2.7  &  1.1 &  1.0   \\\cline{2-8}
&M6 & 3.7  & 3.7 &  38.8 &0.2  & 0.1 &  0.9  \\\cline{2-8}
&M7 & 87.1 & 1.4 &  10.0 & 2.5  & 10.0 &  0.4   \\\cline{2-8}
&M8 & 95.1 & 4.3 &  28.9 &0.8  & 0.4 &  1.2  \\\cline{2-8}
&M9 &  0.0 & 0.5 &  7.5 &4.5  & 0.2 &  1.1   \\\cline{2-8}
&M10 & 5.1 & 9.6 &  11.0 &13.0  & 0.8  &  0.6   \\\cline{2-8}
&M11 & 61.4& 63.2&  27.1 &0.2  & 0.1 &  1.7 \\\cline{2-8}
&M12 & 90.8& 5.4 &  13.1 &0.8  & 0.3 &  1.1 \\\cline{2-8}
&M13 & 17.2& 23.3&  4.33 &0.1  & 0.0 &  0.1 \\\cline{2-8}
&M14 & 27.1& 30.4&  48.5 &0.2  & 0.1 &  1.0 \\\cline{2-8}
&M15 & 0.6 & 0.6 &  51.0 &0.6  & 0.8 &  1.1 \\\cline{2-8}
&M16 & 4.6 & 1.3 &  32.1 &0.2  & 0.4 & 1.0  \\\cline{2-8}
&M17 & 4.9 & 12.6&  32.2 &4.7  &0.0  &  0.0 \\\cline{2-8}
&M18 & 95.2& 13.4& 35.6 &0.2  & 0.8 &  0.4\\\hline\hline
\multirow{8}{*}{\rotatebox[origin=c]{90}{\textbf{Video WA PRNU \cite{r10}}} }& M19 & 17.6& 17.2& 12.1 & 0.5  & 0.5 &  1.8 \\\cline{2-8}
&M20 & 22.3& 30.1& 13.6 & 0.5  &0.4  &  0.8 \\\cline{2-8}
&M21 & 2.8 & 2.4 & 13.8 & 1.0 & 0.8 &  1.3 \\\cline{2-8}
&M22 & 87.6& 87.6& 22.9 & 0.3 &  0.4 &  0.9 \\\cline{2-8}
&M23 & 77.6& 76.8& 14.0 & 3.9 & 3.0  &  1.3 \\\cline{2-8}
&M24 & 3.6 & 2.4 & 12.6 & 20.2 & 17.4 &  1.0 \\\cline{2-8}
&M25 & 48.0& 26.8& 42.2 & 23.0 & 23.7 & 0.7 \\\cline{2-8}
&M26 & 87.6& 84.0& 27.5 & 1.5 & 1.6 &  1.9 \\\hline\hline
&\textbf{Overall} & 41.9 & \underline{28.06} & \textbf{23.4} &3.6 &	\underline{2.8}	& \textbf{0.92} \\\hline
  \end{tabular}
\end{table}

\subsection{Evaluation using the Video-ACID +Video WA PRNU dataset }

The same procedure is used to extract the data from the Video-ACID dataset consisting of comprising of videos from 18 smartphone models. The data is combined with 8 devices from Video WA PRNU dataset to train and test the proposed method. The frames are extracted from each video and cropped as windows with a size of $224\times 224$ each. These windows are cropped from central frames without changing the content of the original frames. 

The evaluation of the proposed method PDC-ViT using the Video-ACID dataset has been computed using different metrics including False Negative Rates (FNRs), and False Positive Rates (FPRs), as well as the confusion metric of the testing set. The performance accuracy of the proposed PDC-ViT method achieved 74.77\%. The results obtained can be shown in the confusion matrix presented as shown in Figure \ref{conf-acid} which demonstrates the ability of PDC-ViT to identify the source camera with minimal error rates.
\begin{figure}[t!]
    \centering
    \begin{subfigure}[b]{0.7\textwidth}
        \includegraphics[width=\textwidth]{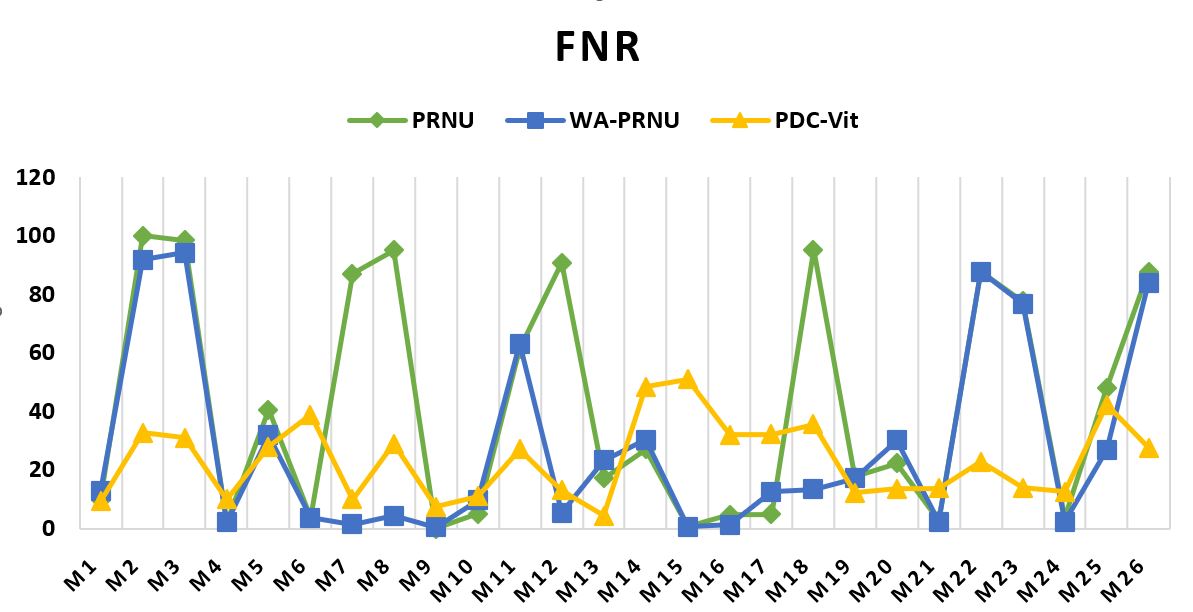}
        \caption{FNR}
    \end{subfigure}\\
    \begin{subfigure}[b]{0.7\textwidth}
        \includegraphics[width=\textwidth]{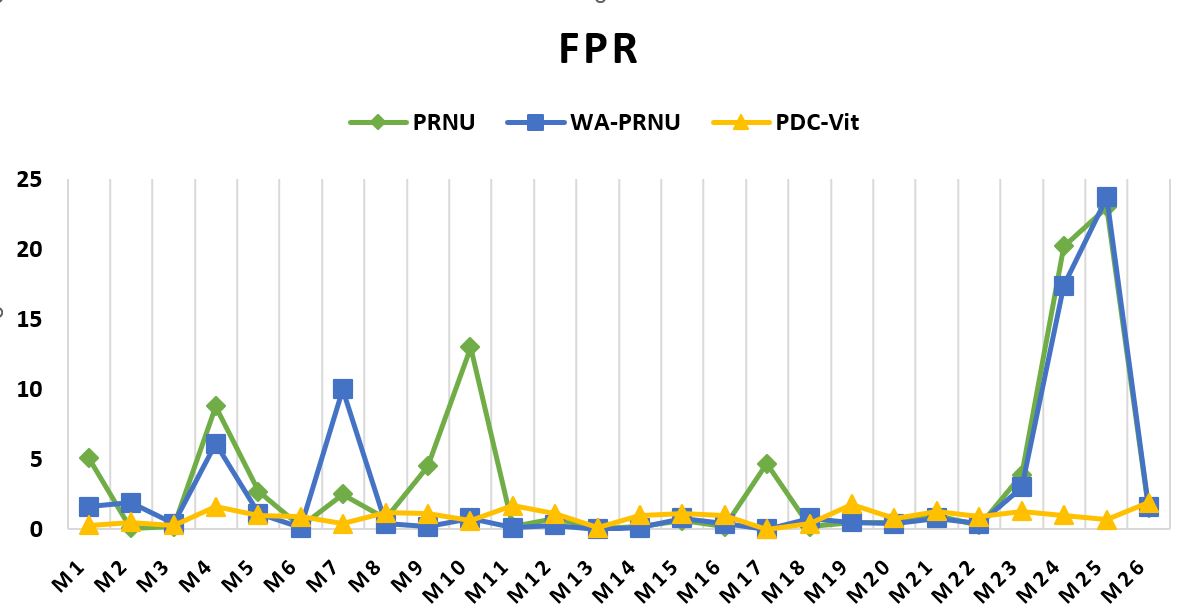}
        \caption{FPR}
    \end{subfigure}  

    \caption{FNR and FPR of of PDC-ViT  and state-of-the-art methods on Video-ACID + Video WA PRNU datasets. }
    \label{fnracid}
\end{figure}
The proposed method is also evaluated using FNTs and FPRs including a comparative analysis against some state-of-the-art methods. The obtained FNRs and FPRs for each smartphone model are presented in Table \ref{table_acid} where the values are visualized in Figure \ref{fnracid}. From the Table, we can observe that the proposed method reached the min FNR and FPR values in terms of overall values, while it can also be seen that the PDC-ViT method reached 23.4\% with a 5\% decrease using Video WA PRNU \cite{r10} and 18.5 using Basic Video PRNU \cite{r10} methods. This is still correct for FPR values, while the proposed methods used frames of $512\times 512$ and we used $224\times 224$ for the PDC-ViT method.

From Figure \ref{fnracid}, the graph demonstrates the capacity of the proposed method to identify the source camera in a stable way compared with the other methods. For example, the FNR values of models from M1 to M4 in the Basic Video PRNU \cite{r10} and Video WA PRNU \cite{r10} methods reached up to 80\% and for the model from M6 to M9 the FNR values are very small. However, the FNR values using the proposed method on the same sets are stable and do not exceed 40\%. The same observation is for FPR values. This demonstrates the high performance of the proposed method.

\section{Challenges and future directions}

False identification is a critical concern in the context of source camera identification, especially considering the advancements of the tools that can manipulate (forge) visual data. Due to this manipulation a method can incorrectly identify an image from a camera that did not capture it. From the manipulation techniques one can find masking by emulators, encryption, or artificial substitution of identifiers, which aim to obscure or alter identifying features. To handle this problem, we have primarily focused on the performance of the PDC-ViT model when distinguishing between different camera sources based on pixel differences that can be presented as a Robust Feature Extraction technique that employs Pixel Difference Convolution to capture subtle pixel-level variations that are inherent to different camera models. These distinctive features are less susceptible to being fully masked or altered by manipulation techniques, reducing the risk of false identification. In addition, the Vision Transformer enhances the model's ability to recognize complex patterns and relationships within images. This strength helps to differentiate source characteristics from those altered by masking or substitution, providing more reliable camera identification. Also the training on a large range of images helps any model including the proposed method to learn from different scenes to reduce the false identification.

The integration of Vision Transformers with Pixel Difference Convolution enhances the accuracy of pixel-level analysis, improving source camera identification. This method's efficiency supports faster processing times and handles large datasets with fewer parameters. While the use of ViT in image classification improved the this including source camera identification, yet some challenges need to be further improved. From these challenges we can mention data availability, computational cost and memory optimization, and Training Instability especially with the improvement of the devices every year.

One major area for future research relates to data efficiency and augmentation. ViTs often require large datasets to perform efficiently, as they lack the inherent inductive biases present in CNNs. Future work should aim to enhance ViTs' data efficiency by developing novel data augmentation strategies that cater specifically to the global self-attention mechanism of ViTs. Techniques like TokenMix have already shown promise in this area, improving ViT performance by enabling models to focus on relevant image areas more effectively \cite{f1}. 

Another critical research direction involves computational and memory optimization. The source camera identification datasets are computationally expensive, and the memory demands are significant. Efforts to mitigate these issues, such as the PatchDropout method \cite{f2}, have made progress in reducing FLOPs and memory requirements. However, more work is needed to optimize these models further for real-time applications. Future research could explore more lightweight models and co-optimization strategies that improve efficiency without compromising accuracy.

Training stability also remains a crucial area of exploration. For example, ViTs are known for their sensitivity to hyperparameters, making them difficult to optimize compared to CNNs. Future research should focus on developing more robust training strategies, such as advanced optimizers or training schemes that reduce the reliance on extensive hyperparameter tuning \cite{f3}. Additionally, addressing overfitting through improved augmentation techniques, like TransMix \cite{f4}, or employing self-attention regularization could lead to more stable and generalizable ViT models.

\section{Conclusion}
This paper proposes a new method for source camera identification. Most of the existing deep learning methods use known backbones to capture general content information from images without considering the pixel representation of the analyzed images. This leads to the misidentification of some source cameras. Our method overcomes this problem by using the concept of Pixel difference convolutions combined with a transformed network. The proposed network extracts the features using the PDC method and then embeds them as the patches’ input of a Vision Transformer network. The proposed system, referred to as PDC-ViT, has been evaluated on five different datasets including Vision, Daxing, Socrates, and QUFVD. The results obtained have demonstrated the superiority of the proposed system over state-of-the-art competing methods. The PDC-ViT network proved to be effective for datasets that contain a wide variety of brands and models, such as the Vision and Socrates datasets. For example, with the Socrates dataset that consists of more than 100 classes representing 100 different devices from 15 brands, the proposed method achieved an accuracy of 94\%. This performance was consistent with the results obtained on the Vision dataset as well. This work not only contributes to the ongoing advancements in source camera identification but also offers practical applications in forensic investigations and digital media verification. Future efforts will focus on further enhancing the model to keep pace with emerging camera technologies and improve adaptability to new devices entering the market.


\section*{Acknowledgements}

This publication was made possible by NPRP grant \#
NPRP12S-0312-190332 from Qatar National Research Fund (a member of Qatar Foundation). The statement made herein are solely the responsibility of the authors.


\section*{Funding}
Authors declare no funding for this research.

\section*{Data availability}
The datasets used are available in internet and public.
\section*{Code Availability}
The code will be published when the paper is accepted.

\section*{Declarations}
\section*{Conflict of interest}
The authors declare that they have no conflict of interest

\end{document}